\documentclass[10pt, a4paper]{article}

\usepackage{lrec-coling2024} 
\usepackage{amsmath}
\usepackage {amssymb}
\usepackage{multirow}
\usepackage{arydshln}
\usepackage{subfigure}

\title{Landmark-Guided Cross-Speaker Lip Reading with Mutual Information Regularization}

\name{Linzhi Wu$^{1,3}$, Xingyu Zhang$^{2}\sthanks{\ \ Corresponding author.}$, Yakun Zhang$^{2,3}$, Changyan Zheng$^{2,4}$,\\\textbf{\large Tiejun Liu$^{1}$, Liang Xie$^{2,3}$, Ye Yan$^{1,2,3}$, Erwei Yin$^{2,3}$}} 

\address{$^{1}$School of Life Science and Technology, University of Electronic Science and Technology of China,\\ Chengdu, China\\
$^{2}$Defense Innovation Institute, Academy of Military Sciences, Beijing, China\\
$^{3}$Tianjin Artificial Intelligence Innovation Center, Tianjin, China\\
$^{4}$High-tech Institute of Qingzhou, Weifang, China\\
         lindgew@std.uestc.edu.cn, zhangxingyu1994@126.com, ykzhang1222@126.com\\
         echoaimaomao@163.com,
         liutiejun@uestc.edu.cn,
         xielnudt@gmail.com\\
         yy\_taiic@163.com,
         yinerwei1985@gmail.com
}

\abstract{
Lip reading, the process of interpreting silent speech from visual lip movements, has gained rising attention for its wide range of realistic applications. Deep learning approaches greatly improve current lip reading systems. However, lip reading in cross-speaker scenarios where the speaker identity changes, poses a challenging problem due to inter-speaker variability. A well-trained lip reading system may perform poorly when handling a brand new speaker. 
To learn a speaker-robust lip reading model, a key insight is to reduce visual variations across speakers, avoiding the model overfitting to specific speakers. In this work, in view of both input visual clues and latent representations based on a hybrid CTC/attention architecture, we propose to exploit the lip landmark-guided fine-grained visual clues instead of frequently-used mouth-cropped images as input features, diminishing speaker-specific appearance characteristics. Furthermore, a max-min mutual information regularization approach is proposed to capture speaker-insensitive latent representations.
Experimental evaluations on public lip reading datasets demonstrate the effectiveness of the proposed approach under the intra-speaker and inter-speaker conditions.
 \\ \newline  }

\begin{document}

\maketitleabstract

\section{Introduction}
Lip reading, commonly known as visual speech recognition (VSR), aims to automatically recognize spoken text units through the speaker's lip movements of a silent video clip, and is widely used in various potential applications such as aiding individuals with hearing impairments
, speech recognition in noisy environments, human-computer interaction \cite{ChungZ16,lrs3ted18,YangZFYWXLSC19,AfourasCSVZ18,RekikBM15a}. Recently lip reading research has made great progress thanks to the advent of deep learning and the availability of large-scale annotated corpus. Particularly, the advanced neural models adapted from the fields of automatic speech recognition (ASR) and natural language processing (NLP) significantly boost the performance of lip reading \cite{AssaelSWF16,ChungSVZ17,Martinez0PP20,0001PP21a,MaPP22}. 

Despite its success, lip reading still suffers from a non-trivial problem for practical usage, namely the considerable variations between speakers \cite{AlmajaiCHL16,BurtonFSNB18}.
Conventional lip reading systems trained on a limited set of speakers tend to recognize the lip movements of specific individuals, and are easily sensitive to speaker variance, making them more suitable for overlapped speakers appeared in the training set. However, different speakers usually have different lip appearance and shapes even when they say the same utterances, and those systems may be prone to overfit the visual variations of lip region, which results in degraded performance when adapting to a speaker never seen before \cite{HuangLF21,LipFormer23}. Hence, it is essential to develop a lip reading system that can be generalized across speakers in favor of real-world applications.

To improve the robustness and accuracy of a lip reading model when dealing with unseen speakers,
one intuitive solution is to eliminate the visual variations across speakers as much as possible. Since facial landmarks are sparse geometric coordinate points, indicating the location of key facial areas, they are robust to the pixel-based visual appearance and could serve as speaker-independent clues \cite{MorroneBPFTB19}. In \cite{LipFormer23}, besides the visual features extracted from lip images, the authors introduced the facial landmarks to suppress the speaker variance in lip shapes and movements, achieving effective performance gains. In addition to the visual clues, another trend is to encourage the lip reading model to learn speaker-independent but speech content related visual representations by various means \cite{WandS17,YangWZZ20,HuangLF21,ZhangWC21,lu2022siamese}.



Existing studies mostly take mouth-centered crops as input, but the visual variations of lip shapes and appearance may be inevitably introduced.
To handle speaker variations, we rethink both the input visual clues and intermediate latent representations in this work. For the visual clues, we make the most of the lip landmarks and explore the landmark-guided fine-grained visual features from three aspects.
First, we consider the landmark-centered patches as they are not only key areas closely related to lip reading, but also facilitate reducing lip shape variance. 
In particular, we extract tubelets (\textit{i.e.}, 3D patches) centered on landmarks in view of both spatial and temporal dimensions.
Second, to build the geometric correlation between different patches within each frame, the relative distances between landmarks are used as positional information to complement the geometric features. Moreover, as lip motion trajectories tend to be speaker-independent, we obtain the lip motion features from landmark tracks across frames by calculating the inter-frame difference of landmark coordinates.
The aforementioned visual features can be regarded as the front-end obtained features, which are then fed into a back-end conformer encoder to model global and local temporal relationships \cite{GulatiQCPZYHWZW20}. Finally, a hybrid CTC/attention architecture \cite{hori-etal-2017-joint,PetridisSMTP18} is utilized for target text prediction. 

Although the fine-grained local visual features induced by lip landmarks are expected to reduce the visual appearance variance, redundant speaker-specific characteristics may still be preserved within some patches. Therefore, we propose to leverage a max-min mutual information (MI) regularization scheme to decouple the identity-related features and speech-related features, facilitating speaker-insensitive latent representations. More specifically, in order to dig out speaker identity-related information, a speaker identification module is additionally introduced. Then, we minimize the MI between the speech-related features extracted by the conformer encoder and identity-related features extracted by the speaker identification module, while maximizing the MI between the representations encoded by the front-end and back-end of a lip reading model.
In conclusion, the major contributions of this paper are summarized as follows:
\begin{itemize}
    \item We investigate the landmark-guided fine-grained visual clues tailored for the cross-speaker lip reading task, with better interpretability in contrast to the widely-used mouth-cropped images.
    \item We propose a max-min mutual information regularization approach to encourage the lip reading model to learn speaker-insensitive latent representations.
    \item Experiments and analysis performed on public sentence-level lip reading datasets demonstrate the effectiveness of the proposed approach in the cross-speaker setting.
\end{itemize}

\section{Methodology}
\begin{figure*}[!t]
\centering
\includegraphics[width=6.3in]{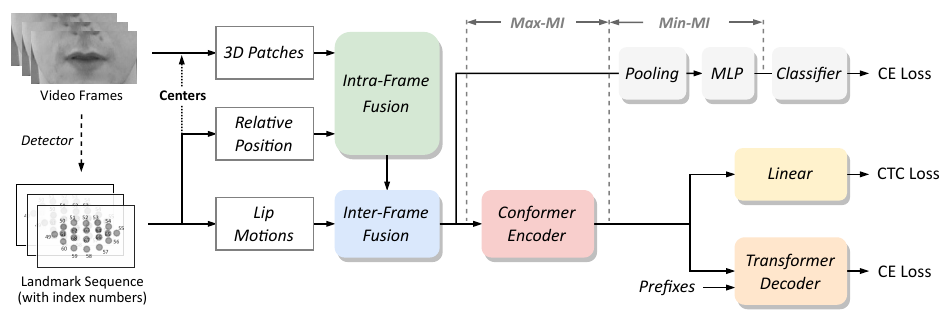}
\caption{Illustration of the overall multi-task learning framework for cross-speaker lip reading. The model inputs are derived from the mouth-centered crops coupled with lip landmarks. 
}
\label{fig:framework}
\end{figure*}

Figure \ref{fig:framework} briefly illustrates the overall model framework built on the joint CTC/attention architecture. It consists of several key components, each of which will be detailed in the following subsections.



\subsection{Model Architecture}
Let $\mathbf{x}=[x_{1}, x_{2},\cdots,x_{T}]$ be the visual input streams of length $T$ drawn from a facial video clip, mapped into the target text sequence $\mathbf{y}=[y_{1}, y_{2},\cdots,y_{N}]$ with $N$ tokens by a lip reading model. Suppose $K$ lip landmarks are detected for each video frame in the pre-processing stage.

\subsubsection{Landmark-Guided Visual Front-end}\label{sec:front_end}
The mouth-centered cropped regions are commonly used as the visual clues. Nevertheless, local subtle lip dynamics (\textit{e.g.}, mouth contours) may fail to to be effectively captured \cite{ShengZXPL22}. 
In addition, the whole mouth-cropped images potentially contain much speaker-related 
information (\textit{e.g.}, personal appearance traits), resulting in more significant visual appearance difference across speakers.
Thus, we investigate the landmark-based visual clues to capture local fine-grained lip movements and meanwhile reduce visual variance among different speakers.
More specifically, we exploit the 2D landmarks of lip region from three distinct perspectives as follows:
\paragraph{3D Patches}
For each frame at time-step $t$, we first extract a small square window (\textit{patch})\footnote{For the sake of simplicity and ensuring a symmetric spatial context, the square-shape window is given preference.} of size $w_t\times w_t$ centered at each landmark position $p_{t}^{i}~  (1\leq i\leq K)$, and the landmark-centered patch describes this point by the surrounding spatial context. Considering the neighboring temporal context of lip movements between adjacent frames, we further extract the 3-dimensional patch (\textit{tubelet}) achieved by a 3D convolutional module, producing the tubelet embedding $\mathbf{v}_{t}^{i}$. In other words, we can construct a tubelet of size $w_t\times w_t\times d$ ($d$ means the depth of patch determined by the kernel size in time dimension) around each lip landmark at each time-step.
Unlike the whole mouth-cropped images encoded by 3D convolution, the landmark-centered tubelet reduces the computational complexity of visual feature extraction. Moreover, less speaker identity-related information is retained.

\paragraph{Intra-Frame Relative Position}
As lip landmarks within a frame are not in fixed and regular position but distributed in a certain shape, the local patches alone may be not sufficient to learn good visual features. Thus, we consider the geometrical relationships between them within a frame by calculating the relative distance between any two landmark points. 
Specifically, we adopt a Multi-Layer Perceptron (MLP) layer to encode the coordinate differences between the $i$-th landmark and other landmarks at the $t$-th time-step, producing the relative positional vector defined as:
\begin{equation}
    \begin{split}
      \mathbf{r}_{t}^{i} = \mathrm{MLP}(\{p_{t}^{i} - p_{t}^{j}\}_{i\neq j}), ~~ (1 \leq j \leq K)     
    \end{split}
\end{equation}

As a result, we have the position-aware visual feature at the $i$-th landmark: $\mathbf{u}_{t}^{i} = \mathbf{v}_{t}^{i} + \mathbf{r}_{t}^{i}$. 
In order to obtain the visual representations of the whole frame, we leverage a attention-weighted aggregation for the fine-grained features of all the landmarks at the $t$-step, allowing for the interaction between those tubelets. Concretely, a $L_{f}$-layer attentive encoder consumes the sequence of tubelet vectors $\mathbf{z}_{t} = [\mathbf{u}_{t}^{1}, \mathbf{u}_{t}^{2},\cdots, \mathbf{u}_{t}^{K}]$ at the $t$-step. Each layer is composed of multi-head self-attention (MHSA) \cite{VaswaniSPUJGKP17} and MLP blocks along with layer normalization (LN) as follows: 
\begin{equation}
    \begin{split}
        \mathbf{y}_{t}^{(l)} &= \mathrm{MHSA}(\mathrm{LN}(\mathbf{z}_{t}^{(l)})) + \mathbf{z}_{t}^{(l)}, \\
        \mathbf{z}_{t}^{(l+1)} &= \mathrm{MLP}(\mathrm{LN}(\mathbf{y}_{t}^{(l)})) + \mathbf{y}_{t}^{(l)}, \\
        (l &= 1\cdots L_{f})
    \end{split}
\end{equation}
The outputs of the last layer ($\mathbf{z}_{t}^{(L_{f})}$) followed by a global average pooling over all the landmarks produce the intra-frame visual features $\mathbf{f}_t$.


\paragraph{Inter-Frame Lip Motions}
We explicitly extract the inter-frame lip movement features derived from the lip landmark tracks. For the $t$-th time-step, we mainly consider the contour and geometric information involving lip dynamics, including the landmark's \textit{x-y} coordinates; the height and width of outer and inner lip measured by Euclidean distance. Because these metrics may vary significantly when a speaker pronouncing. The motion vector can be obtained by computing the difference between two adjacent frames, \textit{i.e.}, the current frame ($t$) simply subtracts the pre-frame ($t-1$). The motion vector of the first time-step can be set to zero. Here we use a 1-dimensional convolutional module to extract context-aware motion features $\mathbf{m}_{t}$.
Furthermore, we combine the motion features with the orthogonal intra-frame visual features to generate the front-end visual representations: 
$\mathbf{h}_{t} = \mathbf{f}_{t} || \mathbf{m}_{t} ~ (t = 1\cdots T)$
 through a simple concatenation operation ($||$).



\subsubsection{Conformer Back-end}
In light of the sequential spatio-temporal properties of video data, the spatially dominant visual front-end mentioned above may fail to capture temporal dependencies between video frames effectively. Hence, we take advantage of the conformer encoder which integrates self-attention mechanisms and convolutional operations, to model global and local temporal dependencies across frames dynamically. 
The front-end visual features are passed through the conformer encoder with $L_{b}$ sequentially stacked blocks with
identical structure: 
\begin{equation}\label{eq:front}
    \begin{split}
        \mathbf{H}^{(0)} &= [\mathbf{h}_{1}, \mathbf{h}_{2},\cdots, \mathbf{h}_{T}], \\ 
        \mathbf{H}^{(l)} &= \mathrm{ConformerBlock}(\mathbf{H}^{(l-1)}),  \\
        (l &= 1\cdots L_{b})
    \end{split}
\end{equation}
Each conformer block with the macaron-like structure is composed of a set of stacked modules: a feed-forward module, a multi-head self-attention module, a convolution module and a second feed-forward module \cite{GulatiQCPZYHWZW20}. 

Similar to ASR, the monotonic alignment property between input and target sequences is also supposed to be satisfied in VSR. To this end, the auxiliary CTC loss over the encoder outputs is applied to maximize the correct target alignments:
\begin{equation}
    \begin{split}
        \mathcal{L}_{CTC} = -\log p_{\mathrm{CTC}}(\mathbf{y}|\mathbf{x}),
    \end{split}
\end{equation}
where $p_{\mathrm{CTC}}(\mathbf{y}|\mathbf{x})\approx \prod_{t=1}^{T} p(y_{t}|\mathbf{x})$ based on conditional independence assumption between the predicted outputs.

\subsubsection{Transformer Decoder}
Autoregressively, a standard transformer decoder \cite{VaswaniSPUJGKP17} is applied to receive the front-end encoded hidden representations and the prefixes of the target sequence to generate the next token of speech content. The decoder is comprised of a embedding layer followed by $L_{d}$ stacked multi-head attention blocks. The sequence of prefixes is projected to embedding vectors, and then the absolute positional encoding is added.
Each attention block consists of a masked multi-head self-attention module, an encoder-decoder multi-head attention module and a feed-forward module.
\begin{equation}
    \begin{split}
        \mathbf{S}^{(0)} &= \mathrm{TE}(\mathbf{\hat{y}}) + \mathrm{PE}(\mathbf{\hat{y}}), \\ 
        \mathbf{S}^{(l)} &= \mathrm{DecoderBlock}(\mathbf{S}^{(l-1)}, \mathbf{H}^{(L_{b})}), \\
        (l &= 1\cdots L_{d})
    \end{split}
\end{equation}
where $\mathbf{\hat{y}}$ is the prefixes of target sequence, $\mathrm{TE}$ and $\mathrm{PE}$ denote the token embedding layer and positional encoding layer respectively.
To generate the desired output sequence, the cross-entropy based training loss is defined to narrow the gap between the predicted sequence and the target sequence.
\begin{equation}
    \begin{split}
        \mathcal{L}_{CE} = -\log p_{\mathrm{CE}}(\mathbf{y}|\mathbf{x}),
    \end{split}
\end{equation}
where $p_{\mathrm{CE}}(\mathbf{y}|\mathbf{x}) \approx\prod_{t=1}^{N} p(y_{t}|y_{<t}, \mathbf{x})$ based on the chain rule during the step-wise decoding process.

Finally, the training objective of the encoder-decoder architecture is calculated by a simple linear combination as follows: 
\begin{equation}
    \mathcal{L}_{VSR} = \lambda \mathcal{L}_{CTC} + (1-\lambda) \mathcal{L}_{CE}.
\end{equation}
where the relative weight $\lambda$ satisfies $0\leq\lambda\leq1$.

\subsection{Speaker Identification}
To allow for the subsequent decoupling of speaker-related features and speech content related features, the model should be able to distinguish different speakers from local lip appearance. Therefore, we introduce an additional speaker identification branch. 
The underlying speaker identity-related features are obtained via a MLP head that has a batch normalization layer, a ReLU activation function and a fully-connected (FC) layer. The speaker classifier with a FC layer is then used for speaker identification based on multi-class cross-entropy objective function. 
\begin{equation}\label{eq:spk}
    \begin{split}
        \mathbf{h}^{ID} &= \mathrm{MLP}(\mathrm{GAP}(\mathbf{H}^{(0)})), \\
        p^{ID} &= \mathrm{softmax}(\mathrm{FC}(\mathbf{h}^{ID})),  \\
        \mathcal{L}_{ID} &= -\sum_{c=0}^{C-1} y_{c}^{ID}\log p_{c}^{ID},
    \end{split}
\end{equation}
where $\mathrm{GAP}$ is the global average pooling layer used to temporally aggregate the front-end visual features (from Eq. \ref{eq:front}), $C$ refers to the number of all speakers, $p_{c}^{ID}$ represents the class probability of the input sample belongs to speaker $c$ while $y_{c}^{ID}$ is the binary ground-truth label indicating whether this sample belongs to the speaker or not. The speaker identification branch is only used in the training phase.

\subsection{Max-Min Mutual Information Regularization}
Although the landmark-based visual clues potentially reduce the inter-speaker visual variations, there may still retain the redundant speaker-specific information within some patches. 
Therefore, we further exploit the speaker-insensitive features in latent representation space. To ensure the independence between speaker identity and speech content, we adopt the mutual information (MI) regularization method to facilitate learning of independent disentangled representations.
Formally, the basic definition of MI between variables $X$ and $Y$ is established on the Kullback-Leibler (KL) divergence of their joint and marginal probability distributions:
\begin{equation}\label{eq:mi}
    \begin{split}
        \mathcal{I}(X, Y) = D_{\mathrm{KL}}\big( p(X, Y)||p(X)p(Y) \big),
    \end{split}
\end{equation}

For differentiable MI estimation, we introduce using the variational Contrastive Log-Ratio Upper Bound (vCLUB) \cite{ChengHDLGC20} as the upper bound of the desired MI $\mathcal{I}(X, Y)$, achieving MI minimization between latent representations.
\begin{equation}
    \begin{split}
        \mathcal{I}_{\mathrm{vCLUB}}(X, Y) := & ~\mathbb{E}_{p(X, Y)}\big[\log q_{\phi}(y|x)\big] - \\
        & ~\mathbb{E}_{p(X)}\mathbb{E}_{p(Y)}\big[\log q_{\phi}(y|x) \big],
    \end{split}
\end{equation}
where the variational distribution $q_{\phi}(y|x)$ with parameter $\phi$ is applied to approximate the unknown conditional distribution $p(y|x)$.\footnote{Note that $\mathcal{I}_{\mathrm{vCLUB}}(x, y)$ remains a MI upper bound when we have a good variational approximation $q_{\phi}(y|x)$ using a neural network.} 
Then, the training loss of minimizing the MI between the content-related features (Eq. \ref{eq:front}) and the identity-related (Eq. \ref{eq:spk}) features can be defined as
$\mathcal{L}_{minMI} = \mathcal{I}_{\mathrm{vCLUB}}(\mathbf{h}^{ID}, \mathbf{H}^{(L_{b})})$.

Besides, we adopt a MI neural estimator based on the Jensen-Shannon divergence \cite{HjelmFLGBTB19} as a lower bound of Eq. \ref{eq:mi} to achieve MI maximization among latent representations. 
\begin{equation}
    \begin{aligned}
        \hat{\mathcal{I}}_{\theta}^{(\mathrm{JSD})}(X, Y) & := \mathbb{E}_{p(X,Y)}\big [-\log(1+e^{-\mathcal{F}_{\theta}(x, y)})\big ] \\
        & - \mathbb{E}_{p(X)p(Y)}\big [\log(1+e^{\mathcal{F}_{\theta}(x, y)})\big ],
    \end{aligned}
\end{equation}
where $\mathcal{F}_{\theta}$ stands for a score function approximated by a MLP with learnable parameter $\theta$.
The front-end is encouraged to capture speaker irrelevant representations, accomplishing by maximizing the MI between features extracted from the front-end and back-end encoders. Thus, the training loss is defined as $\mathcal{L}_{maxMI} = -\hat{\mathcal{I}}_{\theta}^{(\mathrm{JSD})}(\mathbf{H}^{(0)}, \mathbf{H}^{(L_{b})})$ where 
 $\mathbf{H}^{(0)}$ corresponds to the front-end visual features.
 
Consequently, the training objective of the MI estimators is to minimize the overall loss as follows: 
\begin{equation}
    \begin{split}
    \mathcal{L}_{MI} = \mathcal{L}_{minMI} + \mathcal{L}_{maxMI}.
    \end{split}
\end{equation}



\subsection{Training and Decoding}
Finally, to optimize the whole model parameters, we propose to utilize a two-stage training strategy: 
\paragraph{Stage I} We jointly train the VSR module and speaker identification module in a multi-task learning manner, until the speaker identification module converges. 
\begin{equation}
    \begin{split}
        \mathcal{L} = \mathcal{L}_{VSR} + \alpha_{1} \mathcal{L}_{ID}.
    \end{split}
\end{equation}
\paragraph{Stage II} We freeze the weights of the well-trained speaker identification module, and continue to train the VSR module along with the MI estimators.
\begin{equation}
\begin{split}
    \mathcal{L} = \mathcal{L}_{VSR} + \alpha_{2} \mathcal{L}_{MI}.
\end{split}
\end{equation}
where $\alpha_{1}$ and $\alpha_{2}$ are the weight coefficients. 
The first stage is to initialize the speaker identification module, ensuring that speaker identity-related features can be effectively extracted. While the second stage aims to encourage the speech content related features to be towards speaker-invariant. 
During inference, the transformer decoder is performed with a left-to-right beam search algorithm.


\section{Experiments}
\subsection{Datasets and Evaluation}
We conduct experiments on publicly available lip reading dataset GRID \cite{cooke2006audio}.
It is a popular sentence-level dataset, consisting of 34 speakers\footnote{Data source: \url{https://spandh.dcs.shef.ac.uk/gridcorpus}. It is worth noting that the video data for speaker 21 is not available.}, and each speaker utters a set of 1000 sentences with fixed grammar. The duration of each recorded facial video clip is about 3 seconds, sampling 25 frames per second. 
Following \citet{AssaelSWF16}, we utilize the same unseen speaker split that four speakers (1, 2, 20 and 22) are used for testing and the rest for training. For the overlapped speaker setting, we randomly select 255 samples from each speaker for testing and the remaining samples for training. Detailed data statistics can be found in Table \ref{tab:data_stat}.

\begin{table}[!ht]
\centering
\begin{tabular}{cccc}
\hline
\textbf{Setting}   & \textbf{Subset} & \textbf{\#Speaker} & \textbf{\#Sentence} \\ \hline
\multirow{2}{*}{Overlap} & Train  & 33                 & 24408           \\ 
    & Test     & 33      & 8415     \\ \hline
\multirow{2}{*}{Unseen}  & Train     & 29                 & 28837        \\ 
     & Test      & 4    & 3986    \\ \hline
\end{tabular}
\caption{Data statistics for the overlapped and unseen speaker settings. 
}
\label{tab:data_stat}
\end{table}



To measure model performance, we use word error rate (WER) as evaluation protocol following previous literature \cite{AssaelSWF16,ChungSVZ17}. WER in percentage is calculated by comparing the number of substitutions (S), deletions (D), and insertions (I) required to transform the recognized output generated by a lip reading system into the reference transcription, divided by the total number of words in the reference transcription (N). Mathematically, the formula for calculating WER can be just defined as {WER = (S + D + I) / N}. Lower WER values indicate higher accuracy. 


\subsection{Implementation Details}
For the pre-processing, we use the face alignment detector \cite{bulat2017far} to detect and track 68 facial landmarks for each frame of video clips from the dataset\footnote{Open-source toolkit: \url{https://github.com/1adrianb/face-alignment}. According to the same index numbers in consecutive frames, we can connect the corresponding landmark coordinates across frames and track those landmarks over time.}. We resize the original video into 360$\times$288, and select all 20 lip landmarks aligned with the landmark point of the nose tip ($K\!=\!20$). The basic size of each patch centered on a landmark point is 24$\times$24. Moreover, all video frames are converted to grayscale and normalized by  division by 255. 

Table \ref{tab:3dpatch} shows the architecture of the 3D patch encoding module. For the conformer encoder, we use 3 blocks, hidden dim of 256, feed forward dim of 1024, 8 attention heads, and the kernel size of each depth-wise convolutional layer is set to 31.
For the transformer decoder, the basic hyper-parameters are the same as in the conformer ($L_{f}\!=\!L_{b}\!=\!L_{d}\!=\!3$).
$\lambda$ is set to 0.1 as suggested in \cite{MaPP22}, and the values of $\alpha_{1}, \alpha_{2}$ are empirically set to 0.2.
During training, the Adam optimizer 
is used to update the learnable model parameters with a mini-batch size of 50.
The initial learning rate is $3e^{-4}$, following a schedule strategy that increases linearly from 0 to the initial value and thereafter decreases with cosine annealing. 
In the testing phase, a beam search decoder is applied to the transformer decoder for character-level prediction with beam width 10, without using external language model.


\begin{table}
    \centering
    \begin{tabular}{l|c|c}
        \hline
        \textbf{Layers} & \textbf{Filters} & \textbf{Output size} \\ \hline
        Conv3D & $5\times3\times3,64$  &  $64\times T\times\frac{H}{2}\times\frac{W}{2}$ \\
        MaxPool3D & $1\times3\times3,64$  & $64\times T\times\frac{H}{4}\times\frac{W}{4}$ \\
        Conv2D & $3\times3,128$  & $T\times 128\times\frac{H}{8}\times\frac{W}{8}$ \\
        Conv2D & $3\times3,256$  & $T\times 256\times\frac{H}{16}\times\frac{W}{16}$ \\
        AvgPooling & -- & $T\times 256$  \\ 
        \hline
    \end{tabular}
    \caption{The 3D patch encoding module. Each convolution layer is followed by batch normalization and Swish activation function.}
    \label{tab:3dpatch}
\end{table}

\paragraph{Flexible Patch Size} Compared with a fixed patch size, \citet{BeyerI0CKZMTAP23} have demonstrated the superiority of randomized
patch sizes for a standard vision transformer \cite{DosovitskiyB0WZ21}. Drawing inspiration from this, we try to dynamically change the size of landmark-centered patch at each iteration during training, which is implemented by randomly sampling a window size from a range of windows (e.g., $w\times w$, $w\in \{20, 22, 24, 26, 28, 30, 32\}$ with an interval of 2 pixels in this work).\footnote{Due to limited computational resources, we uniformly resize the large cropped patches to a proper resolution of 24$\times$24, which is also used for model inference.}

\subsection{Ablation Analysis}
We perform a series of ablation studies in the unseen speaker setting to better understand our method from different aspects. Results are shown in Table \ref{tab:ablation}. First, performance drop can be observed when removing the relative position or lip motion features from the fine-grained visual clues of our pipeline. That verifies the importance of each part to enhance the visual features. Moreover, we ignore the temporal context information of the landmark-centered patch through replacing the tubelet with 2D patch, resulting in about 1.4\% WER increase. Furthermore, we abandon the mutual information regularization terms from the framework, leading to consistent performance drops.

Here we consider adopting the commonly-used mouth-cropped images rather than the proposed fine-grained visual cues, and the results with performance degradation prove the benefits of our visual cues in improving the recognition performance of unseen speakers. 
The whole mouth regions and beyond may provide more complete and informative spatial context cues beneficial for lip reading \cite{ZhangYXSC20}, but at the cost of preserving more speaker-specific characteristics that are not conducive to cross-speaker adaptation. Also, this motivates us to make a good trade-off between recognition accuracy and robustness when exploiting the visual features.

\begin{table}[!ht]
\centering
\begin{tabular}{l|c}
\hline
\textbf{Method} & \textbf{WER} (\%) \\
\hline
Ours & 10.21 \\ \hdashline
~ w/o \textit{RelPos} & 10.69  \\
~ w/o \textit{Motion} & 10.92  \\
~ w/o \textit{MI} & 11.13  \\\hdashline
~ w/o \textit{RelPos}\&\textit{Motion} & 11.41  \\
~ w/o \textit{RelPos}\&\textit{MI} & 11.65  \\
~ w/o \textit{Motion}\&\textit{MI} & 11.77  \\
~ w/o \textit{RelPos}\&\textit{Motion}\&\textit{MI} & 12.40  \\ \hdashline
Replacing 3D patch with 2D patch & 11.62  \\
Using mouth-centered crops & 11.50  \\
\hline
\end{tabular}
\caption{Ablation studies for unseen speakers. \textit{RelPos}, \textit{Motion} and \textit{MI} mean relative positions among intra-frame landmarks, lip motion information, and mutual information regularization, respectively. \label{tab:ablation}}
\end{table}

\subsection{Comparison with Previous Methods}
As shown in Table \ref{tab:main_res}, we compare the proposed method with the previous competitive baselines for the overlapped and unseen speaker settings.
We can observe that different lip reading methods indeed perform much better when handling those seen speakers.
The comparison results also indicate that the proposed method can achieve performance on par with or exceeding those competitive methods in both settings, by attaining 1.83\% WER and 10.21\% WER in the seen and unseen speaker scenarios respectively. Since the core motivation of this work is not to pursue a new state-of-the-art, here we do not consider the mouth-cropped images like the previous methods as global visual cues having rich spatial information. Improved recognition performance may be further obtained through any feasible integration \cite{ShengZXPL22, LipFormer23} with the proposed fine-grained visual cues.



\begin{table}[!ht]
\centering
\begin{tabular}{l|c}
\hline
\textbf{Method ~(Overlapped)} & \textbf{WER} (\%) \\
\hline
LipNet \cite{AssaelSWF16} & 4.80 \\
WAS \cite{ChungSVZ17} & 3.00 \\
LCANet \cite{XuLCW18} & 2.90 \\
DualLip \cite{Chen0XQWL20} & 2.71 \\
CALLip \cite{HuangLF21} & 2.48 \\
LCSNet \cite{XueYLHCGH23} & 2.30 \\
\hdashline
Ours & \textbf{1.83} \\
\hline\hline
\textbf{Method ~ (Unseen)} & \textbf{WER} (\%) \\
\hline
LipNet \cite{AssaelSWF16} & 14.2 \\
WAS \cite{ChungSVZ17} & 14.6 \\
TM-seq2seq \cite{AfourasCSVZ18} & 11.7 \\
Motion\&Content \cite{Riva0S20} & 19.8 \\
PCPG-seq2seq \cite{LuoYSC20} & 12.3  \\
LCSNet \cite{XueYLHCGH23} & 11.6 \\
\hdashline
Ours & \textbf{10.21} \\
\hline
\end{tabular}
\caption{Performance comparison with previous competitive baseline methods.\label{tab:main_res}}
\end{table}


\subsection{Effect of Patch Size}
\begin{figure}[!ht]
\begin{center}
\subfigure[Overlapped Speaker]{
    \includegraphics[width=0.235\textwidth]{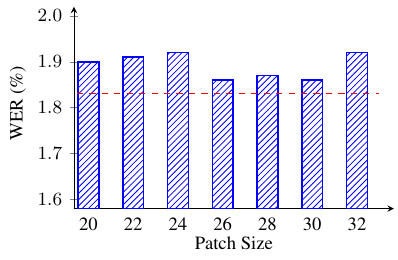} 
}
\hspace{-13pt}
\subfigure[Unseen Speaker]{
    \includegraphics[width=0.235\textwidth]{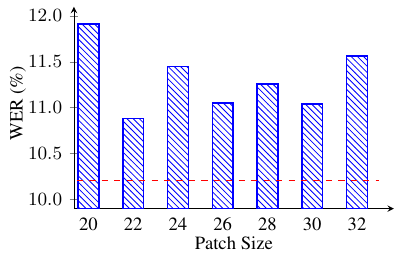} 
}
\caption{Performance comparison of different patch size (ranged from 20 to 32) in the overlapped and unseen speaker settings. The dashed line indicates the recognition performance using flexible patch size.}
\label{fig:unseen_ps}
\end{center}
\end{figure}

To analyze the effect of patch size on recognition performance, we further examine the landmark-centered patches with different sizes, as depicted in Fig. \ref{fig:unseen_ps}.
Large patch size means more spatial contextual information around a landmark point, and vice versa. 
We observe that: (1) Patch size has a less impact on the performance of overlapped speakers compared to unseen speakers. It may be attributed to a fact that the local appearance around a landmark point is similar for a speaker who has already been seen. (2) For unseen speaker setting, smaller patches may provide less spatial contexts, while larger patches may lead to redundant visual information since lip landmarks are closely arranged. Thus, a moderate patch size ensures good recognition results for unseen speakers.

Instead of using a single patch size, we propose to utilize flexible patch size (FPS) for the lip-reading model training by patch sampling strategy. The comparison demonstrates that FPS can lead to better recognition performance (\textit{i.e.}, dashed lines), especially for the unseen speaker setting. 
FPS can actually be regarded as a spatial context augmentation, and a lip reading model trained at FPS enables receive patches of variable scales in comparison to that trained at a single fixed patch size.

\subsection{Attention Visualization}
\begin{figure*}[!ht]
\begin{center}
\includegraphics[width=1\textwidth]{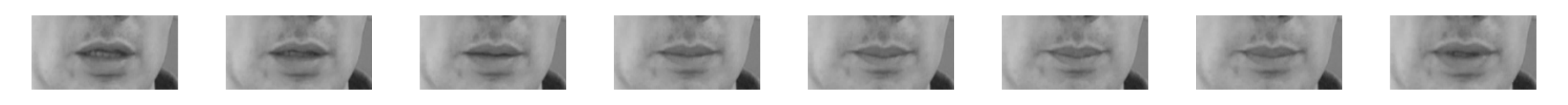} 
\includegraphics[width=1\textwidth]{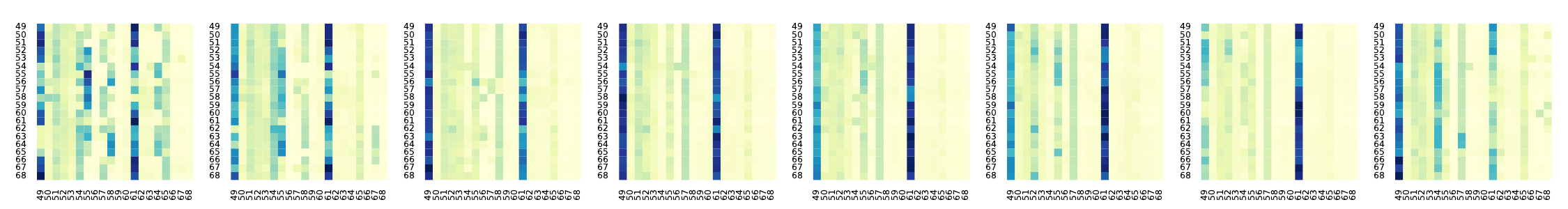} 
\caption{Attention weight maps between lip landmarks (indices from 49 to 68) from the attentive intra-frame fusion module. The weights are calculated by averaging over all the self-attention heads. The video clip used here is drawn from the test set. Darker colors indicate larger weight values.}
\label{fig:attn_map}
\end{center}
\end{figure*}

As mentioned in subsection \ref{sec:front_end},
a multi-head attentive fusion module is used to aggregate the features of 3D patches centered on lip landmarks within a frame. We examine all 20 lip landmarks out of 68 facial landmarks, indexed from 49 to 68. Figure \ref{fig:attn_map} presents the attention maps of a sampled video clip produced by the attentive fusion module. The weights are calculated by averaging over all the self-attention heads at all layers, with values suggesting the importance between the landmarks. We can observe that not all landmark-centered areas are non-trivial across frames, and the module pays more attention to the areas around the corners of the mouth (\textit{e.g.}, landmark 49, 54, 55, 57, 61, etc). One possible reason is that the visual movements of those areas are relatively more evident at a local scale (\textit{"view"}), when a speaker utters with his mouth open and closed. This finding may help us determine lip landmarks that need to be processed, leading to reduced computation overhead.




\section{Related Work}
\subsection{Lip Reading}
Lip reading technique is essentially the translation of lip movements related visual signals into corresponding transcribed text. For the visual signal input, aligned mouth regions of interest cropped by detected landmarks are the most commonly-used \cite{ChungZ16,ChungSVZ17,AfourasCSVZ18,PetridisSMTP18,0001PP21a,MaPP22,LipFormer23}. Using extraoral regions (\textit{e.g.}, the upper face and cheeks) also helps boost recognition performance \cite{ZhangYXSC20}. For the textual output, lip reading paradigms can be broadly categorized as word-level setting and sentence-level setting.
The former aims to map a video frame sequence into isolated units with limited number (\textit{e.g.}, digits, letters or words), which is extensively explored by early research efforts \cite{ChungZ16,YangZFYWXLSC19,Martinez0PP20,ZhaoYSC20}. The latter is challenging yet practical, mapping a video frame sequence into a spoken sentence. Typically, the model architecture of the two paradigms consists of 3D and 2D convolutional layers (\textit{e.g.}, ResNet \cite{HeResNet16}) as the front-end, and sequential models as the back-end, such as RNN \cite{AssaelSWF16,ChungSVZ17}, TCN \cite{Martinez0PP20} and Transformer \cite{AfourasCSVZ18,AfourasCZ18a}. Unlike the word-level trained with simple classification loss, sentence-level lip reading models usually train with Connectionist Temporal Classification (CTC) \cite{GravesFGS06} or sequence-to-sequence \cite{SutskeverVL14,VaswaniSPUJGKP17} fashion to achieve effective performance.

Even with the great progress, existing lip reading systems are restricted in the limited number of speaker, leading to speaker dependency problems. Due to the visual variations of lip movements across speakers, these systems enable achieve considerable performance for overlapped speakers in training set, whereas obtain significant performance drops for unseen speakers. To eliminate the variations, on the one hand, enhancing the input visual clues may be a straightforward way. \citet{Riva0S20} used the motion dynamics derived from simple adjacent-frame differences to improve the performance. \citet{LipFormer23} used facial landmarks to complement the features extracted from lip images. On the other hand, several previous studies instead pay more attention to learning speaker-independent features that are robust to speaker identity information, such as adversarial training \cite{WandS17}, disentangled representation learning \cite{ZhangWC21,lu2022siamese}, and speaker normalization \cite{YangWZZ20,HuangLF21}.

In this work, we focus on the sentence-level lip reading. Unlike the previous studies, considering that the mouth-centered crops might preserve more speaker-related features (\textit{e.g.}, the beard or mole around mouth) irrelevant to speech content recognition, we explore the landmark-guided fine-grained visual clues to reduce visual appearance variance. Moreover, a mutual information regularization scheme is proposed to encourage both the front- and back-end of a lip reading model to learn speaker-insensitive latent representations.

\subsection{Landmark-based Visual Features}
Facial landmarks, referring to specific coordinate points on a person's face that are used to locate key facial areas (\textit{e.g.}, eyes, eyebrows, nose and lip), have attracted increasing attention in visual speech-related fields over recent years. 
One advantage of landmark points is that they outline the overall shape of facial key areas in a sparse positional encoding way, and further geometric and contour features can be easily derived \cite{cetingul2006discriminative,kumar2007profile,zhou2011towards}, effectively describing lip motion irrespective of speakers.
\citet{MorroneBPFTB19} applied motion features based on facial landmarks to improve speech enhancement in a multi-speaker scenario, and proved that the advantages of motion-based features over position-based features. To fully exploit the characteristics of lip dynamics, \citet{ShengZXPL22} leveraged Graph Convolution Network (GCN) to model dynamic mouth contours and capture local subtle movements, improving recognition performance by enhancing visual feature representations. Since landmark-based features are less affected by visual variations caused by lip shapes and appearance, \citet{LipFormer23} introduced the facial landmarks as complementary feature to the visual appearance of lip regions via a cross-modal fusion manner, eliminating biased visual variations between speakers and yield improved performance and robustness for unseen speakers.
Motivated by the success of landmarks,
in this work, we further investigate the lip-landmark guided visual clues 
for facilitating generalization to unseen speakers.

\subsection{Mutual Information Regularization}
Mutual information (MI) is typically employed as a measure of the amount of information that one random variable reveals about the other \cite{kinney2014equitability}. It quantifies the dependence between two variables. 
In the context of (unsupervised) representation learning, through maximizing the MI, the model is enforced to capture meaningful dependencies or relevance between different feature representations, and vice versa. Actually, MI is hard to exactly calculate in the high-dimensional and continuous cases. Thus, various efficient neural estimation methods have been proposed over recent years as approximate solutions \cite{MINE18,InfoNCE2018,HjelmFLGBTB19,ChengHDLGC20}.
\citet{KrishnaB019} improved the image-to-question generation model by maximizing the MI between the image, expected answer, and generated question.
\citet{abs-1812-06589} tried to solve talking face generation generation problem by MI maximization between word distribution and other modal distribution. Similarly, to improve the lip reading performance, \citet{ZhaoYSC20} utilized the global and local MI maximization constraints to extract discriminative features. 

Different from previous works, we introduce a MI regularization term to learn informative representations for cross-speaker adaptation in lip reading. 
Instead of relying on sample pairs from the same or different speakers as model input \cite{YangWZZ20,ZhangWC21,lu2022siamese}, we try to minimize the MI between speaker-dependent features and content-dependent features for the purpose of decoupling, while maximizing the MI between the front-end encoded features and the back-end encoded features.

\section{Conclusion}
In this paper, we provide insights into the cross-speaker lip reading task in terms of visual clues and latent representations, aiming to reduce visual appearance variations across speakers. On the basis of the hybrid CTC/attention architecture, we propose to exploit the landmark-guided fine-grained visual clues as model input features, while introducing the max-min mutual information regularization to learn speaker-insensitive latent representations via a two-stage optimizing scheme. The experimental results evaluated on the sentence-level lip reading demonstrate the effectiveness of the proposed approach. 

\section{Limitations}
One potential drawback of softmax activation in the speaker identification module is that it fails to encourage cluster compactness and cannot ensure the similarity among samples within the same category.
To address this problem, the AM-Softmax loss \cite{wang2018additive}, an enhanced version of softmax, may help better learn speaker discriminative representations.
In addition, the performance of the current lip reading system still has room for further advancement. One possible way to improve is to make the most of the mouth-cropped images and beyond as complementary information. We leave these for future research.

\section{Acknowledgements}
This work was supported in part by the grants from the National Natural Science Foundation of China under Grant 62332019 and 62076250, the National Key Research and Development Program of China (2023YFF1203900, 2023YFF1203903).

\nocite{*}
\section{References}\label{sec:reference}

\bibliographystyle{lrec-coling2024-natbib}
\bibliography{lrec-coling2024}


\end{document}